# An Agentic Model Context Protocol Framework for Medical Concept Standardization

Jaerong Ahn[1], Andrew Wen[1], Nan Wang[1], Heling Jia[2], Zhiyi Yue[1], Sunyang Fu[1] and Hongfang Liu[1]

1 McWilliams School of Biomedical Informatics, The University of Texas Health Science Center at Houston, TX, USA; 2 Department of Artificial Intelligence and Informatics, Mayo Clinic, Rochester, MN, USA.

## Abstract

The Observational Medical Outcomes Partnership (OMOP) common data model (CDM) provides a standardized representation of heterogeneous health data to support large-scale, multi-institutional research. One critical step in data standardization using OMOP CDM is the mapping of source medical terms to OMOP standard concepts, a procedure that is resource-intensive and error-prone. While large language models (LLMs) have the potential to facilitate this process, their tendency toward hallucination makes them unsuitable for clinical deployment without training and expert validation. Here, we developed a zero-training, hallucination-preventive mapping system based on the Model Context Protocol (MCP), a standardized and secure framework allowing LLMs to interact with external resources and tools. The system enables explainable mapping and significantly improves efficiency and accuracy with minimal effort. It provides real-time vocabulary lookups and structured reasoning outputs suitable for immediate use in both exploratory and production environments.

# Introduction

Leveraging observational health data from sources such as electronic health records (EHRs), insurance and administrative claims, and clinical registries has become a cornerstone for generating real-world evidence to advance clinical knowledge and inform medical practice.[1–3] Despite their potential, these data sources are inherently heterogeneous, fragmented, and collected primarily for administrative or clinical purposes rather than research, creating substantial barriers to secondary use. To overcome these challenges, the Observational Medical Outcomes Partnership (OMOP) common data model (CDM) was established to harmonize disparate datasets into a standardized structure and controlled vocabulary, thereby enabling reproducible, large-scale, multi-center research.[4,5]

A critical component of OMOP-based standardization is the mapping of source medical terms to OMOP standard concepts. However, terminology mapping represents a critical bottleneck in clinical data analysis and research. It is resource-intensive, requiring significant manual effort by domain experts, and is prone to inconsistency across sites and studies.[6,7] The challenge arises in part from the flexibility and variability of human language: the same medical concept may appear in multiple forms due to limited terminology standard adoption, local naming conventions, typographical errors, abbreviations, or the use of non-standard coding systems. Automated mapping approaches frequently struggle to resolve such variations with semantic precision, requiring the need for expert validation to ensure accuracy, interoperability, and reliability of downstream analyses.[8]

The advancement in large language models (LLMs) presents opportunities to automate this mapping process but their application in healthcare contexts faces challenges. Recent studies demonstrate that standalone LLMs often generate factually incorrect outputs or fabricate non-existent medical codes, raising concerns about their reliability for the task.[9] These concerns are compounded by the probabilistic, non-transparent nature of LLMs and the high-stake consequences of incorrect mapping of clinical terms. As a result, systematic validation and robust error-mitigation are essential prerequisites before such models can be safely integrated into the analytics workflow.[9]

To address these limitations, we developed a zero-training, hallucination-preventive mapping system based on the Model Context Protocol (MCP), a standardized and secure framework allowing LLMs to interact with external resources and tools.[10] Adopting MCP equips LLM agents with real-time access to the OHDSI Athena, the OMOP vocabulary service. The system enables users to interact with an LLM via natural language prompts to trigger agentic workflow. It interprets clinical terminology, retrieves candidate concepts from Athena, and produces auditable OMOP concept mappings under user guidance.

Unlike prior systems require fine-tuning or pre-trained embeddings and operate on static knowledge bases with deployment complexity,[11–14] our architecture integrates up-to-date vocabulary tools within an agentic MCP framework. This design provides always-current terminology, user oversight, and immediate deployment, therefore enhances reliability, transparency, and usability without sacrificing control or accuracy.

In the following, we present the architecture of the system, report its performance across multiple clinical domains, and discuss its practical value in implementing OMOP CDM transformations.

# Methods

## MCP Architecture

The MCP, recently proposed by Anthropic, provides a standardized and secure framework allowing LLMs to interact with external resources and tools.[10] The proposed architecture consists of a two-step reasoning process, and tool-calling under the guidance of contextual resources provided by the MCP (Figure 1). The MCP server was implemented using the FastMCP framework.[15] When a user sends a query to the LLM client (Azure OpenAI GPT-4o or Claude Sonnet 4), the system analyzes the user prompt to infer a medical term, the target OMOP table and field while interpreting any context requirement. For example, when processing "CP" in the user prompt, the LLM interprets this as "chest pain" based on clinical context and OMOP table specifications. The LLM agent then invokes a pre-defined function to call the Athena OHDSI API search endpoint with the inferred keyword, which presents multiple candidate concepts from the Athena server. The agent then performs a second reasoning process to make a final selection based on semantic closeness, OMOP best practices, and user-specified requirements. In this case, the system selects the most appropriate concept (such as concept ID 77670 for "chest pain") from multiple candidates using OMOP best practices and vocabulary preferences as specified by the MCP resources (Figure 1). These resources provide contextual guidance including OMOP data model specifications, documentation, and vocabulary preferences to ensure clinically appropriate and standardized mappings.

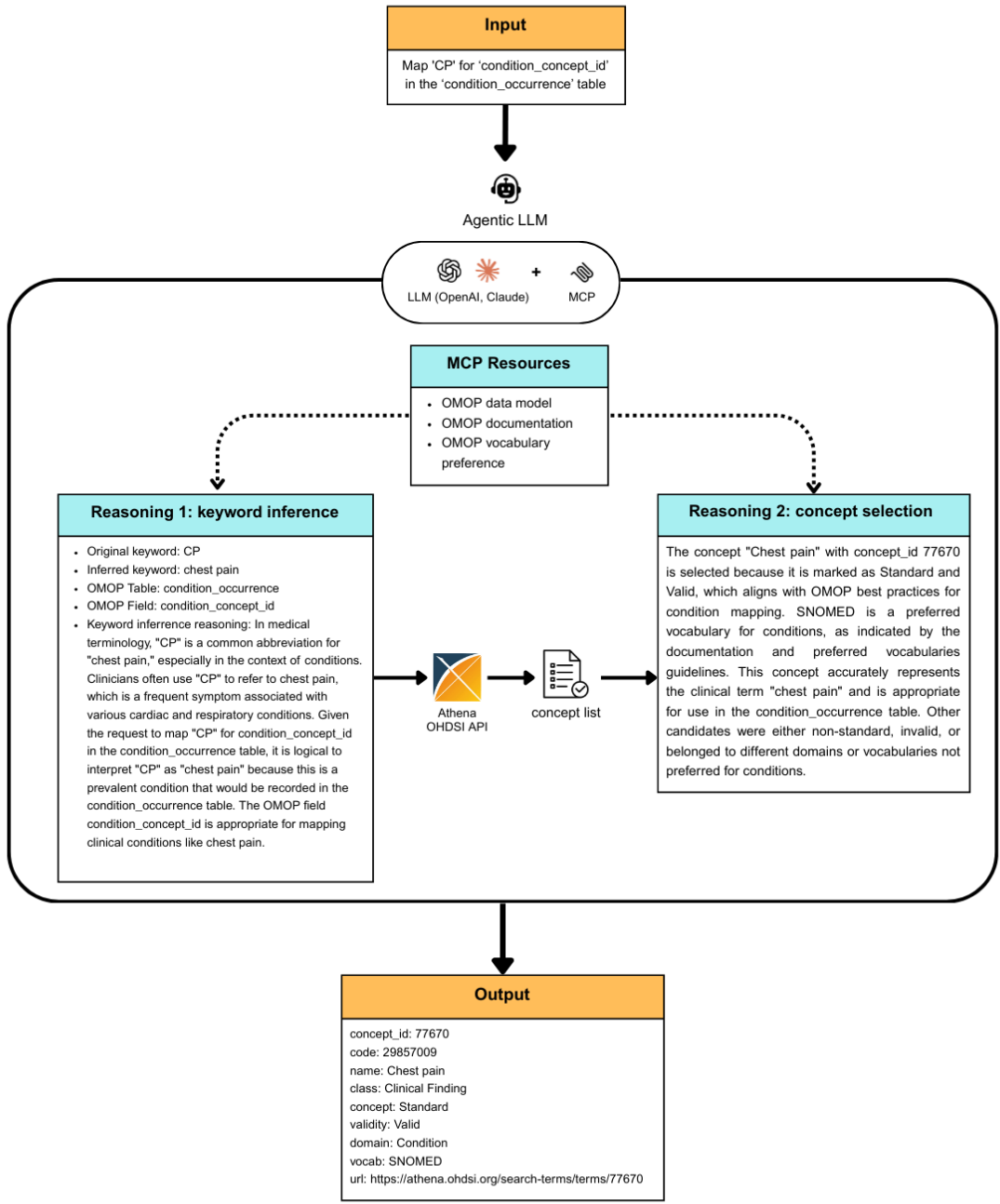

Figure 1. A schematic diagram of the MCP-based concept mapping workflow. Example showing two-step reasoning process: keyword inference ("CP" → "chest pain") and concept selection, guided by MCP resources containing OMOP specifications and vocabulary preferences.

## Prompt Design

The system's prompt is designed to guide the LLM through a structured, multi-step concept mapping process. The prompt begins by explicitly instructing the model to first interpret the user input (e.g., a raw clinical expression), then call the available tool to query the OHDSI Athena via their RESTful API for vocabulary candidates, and finally choose the best match based on semantic fit and metadata context.

To reduce hallucination risk, the prompt employs the following safety strategies:

(a) Explicit tool use requirement: The model is not allowed to invent concept IDs and must use the tool to look them up.

(b) Output format constraints: The model is required to return results in a predefined JSON format containing concept_id, concept_name, domain_id, class, validity, domain, vocabulary, concept URL, and the reasoning for choosing a particular concept.

(c) Custom instructions: By default, the tool was designed to prefer standard, and valid concept, and prioritize domain-specific vocabularies. Specifically, the prompt incorporates soft-coded vocabulary preferences aligned with OMOP recommendations (e.g., SNOMED for conditions, RxNorm for drugs, LOINC for measurements, etc.). Yet, this can be overridden by a user input at runtime. Expert users may direct the model to use non-standard or domain-specific vocabularies (e.g., CPT, ICD-9, etc.) when appropriate, supporting use cases aligned with their research requirement (e.g., legacy data mapping)

(d) Reasoning requirements: To make the selection process more interpretable, the prompt also required the LLM to perform detailed reasoning for keyword inference based on the user prompt and concept selection logic.

## Performance evaluation

The tool was evaluated on two separate datasets to assess (1) the fundamental necessity of the MCP architecture, and (2) utility and generalizability the tool (Table 1). The performance was measured using two primary metrics: (1) retrieval success, which measures the system's basic ability to identify existing, valid concept IDs and names regardless of clinical appropriateness, and (2) relevance score, which measures mapping success on a 0-2 scale. This accounts for the fact that there can be potentially multiple valid mappings for a single term depending on the user requirements.

| Metric | Definition | Measurement | Purpose |
|--------|-----------|-------------|---------|
| Retrieval Success | Percentage of keywords for which a concept ID is successfully identified regardless of relevancy. | Binary (success or failure) | Assess basic mapping ability |
| Relevance Score (0-2 Scale) | Clinical appropriateness of mappings | 0 = Completely wrong<br>1 = Reasonable/usable<br>2 = Optimal | Assess real-world clinical utility |

Table 1. Evaluation metric used for the study.

## Retrieval Failures

A retrieval failure occurs when either a human expert or the LLM system cannot successfully identify a valid concept ID for a given keyword. We categorized three types of possible retrieval failures: (1) no mapping found - the labeler could not identify any appropriate mapping for the given concept, (2) non-existent concept ID - the system produced a concept ID that does not exist in the OMOP vocabulary, and (3) concept ID-name mismatch - the produced concept ID exists but does not correspond to the expected concept name (Table 1).

## Evaluation 1: MCP Server Necessity Assessment

We used 48 medical terms for this assessment. In the No-MCP condition, access to the MCP server was disabled, restricting the LLM solely to its foundational knowledge to solve the task. Both conditions maintained identical input prompts and output format requirements to ensure methodological consistency.

## Evaluation 2: Cross-Domain Performance Evaluation

To assess the generalizability of this tool across different clinical domains, we conducted an expanded evaluation using 150 medical terms randomly curated from UTHealth's OMOP instance. The keywords were selected from three clinical domains: measurement, procedure, and medication. These concepts were already mapped by human experts and deployed in the operational database, providing gold standard labeling set for the evaluation. The keyword sets were randomly split in half and assigned to two independent medical expert evaluators, who were different from the experts that performed the original mappings. They assessed both the human-generated and LLM-generated mappings using the metrics provided in Table 1. If there are multiple valid concept IDs mapped to a single keyword by human experts, the evaluators rated all of them, and the highest scoring mapping was selected for comparison against the LLM's single optimal output.

# Results

## MCP effectively eliminates hallucinations in medical concept mapping

We first evaluated the performance of the concept mapping system in the presence or absence of the MCP server. For this evaluation, we used a curated set of 48 medication terms as medication concepts tend to be less variable, making them suitable for testing baseline performance of the system. We compared the retrieval success and processing time (Figure 2).

The comparison revealed substantial performance differences between two conditions. The MCP-integrated system achieved perfect retrieval success (100%, 48/48 samples) with an average processing time of 5.49s per query. The no-MCP system by contrast, failed to achieve any successful retrievals (0%). The average processing time per query was nearly half (2.13s) (Figure 2A and B), indicating the added overhead of the reasoning and tool calling of the MCP server. In the no-MCP condition, 10.4% of cases produced non-existing concept IDs, and in the remaining 89.6% of cases where valid concept IDs were generated, the concept names and IDs did not match (Figure 2C). The results strongly demonstrate without a clear contextual guidance and external reference sources, LLMs are prone to hallucination.

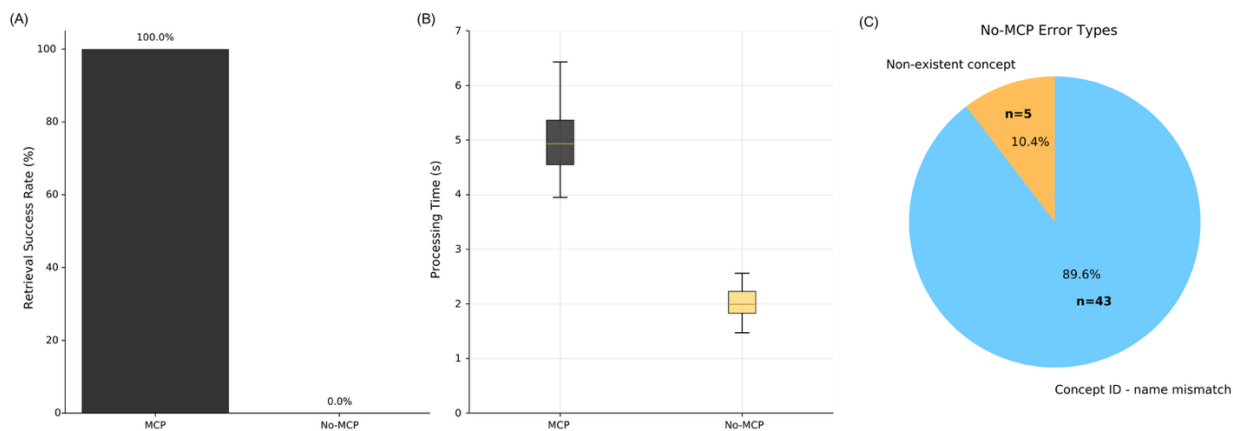

Figure 2. LLM mapping performance comparison with or without the MCP server.

## Multi-domain performance evaluation

To evaluate the generalizability of the MCP system across multiple clinical domains, we tested the framework on 150 terms curated from three OMOP domains (**Methods**). The MCP system significantly outperformed human experts across all evaluation metrics (Figure 3). The system achieved perfect retrieval success (100%, 150/150 keywords) compared to 94.7% (142/150) by human experts, successfully identifying valid mappings for 8 additional terms that human experts were unable to map (Figure 3A).

Among the 142 terms where both methods achieved successful retrieval, we compared the relevance scores. Again, the MCP-aided LLM performed significantly better than human experts in clinical relevance scoring, achieving an average relevance score of 1.61 compared to 1.39 for human experts (p = 0.0073, r = 0.572, Wilcoxon signed-rank test). The score agreement matrix revealed overall superiority of the MCP system in producing higher quality mappings (Figure 3B). Among the 142 concepts successfully retrieved, the MCP system generated fewer completely inappropriate mappings (score 0) in 8.5% of the cases (12/142) compared to 24.6% (35/142) by human experts, while producing optimal mappings in 69.7% of cases (99/142) compared to 64.1% (91/142) by human experts. The system produced the mapping results with an average time of 6.20 ± 0.10 seconds per term, demonstrating both higher accuracy and efficiency in its performance.

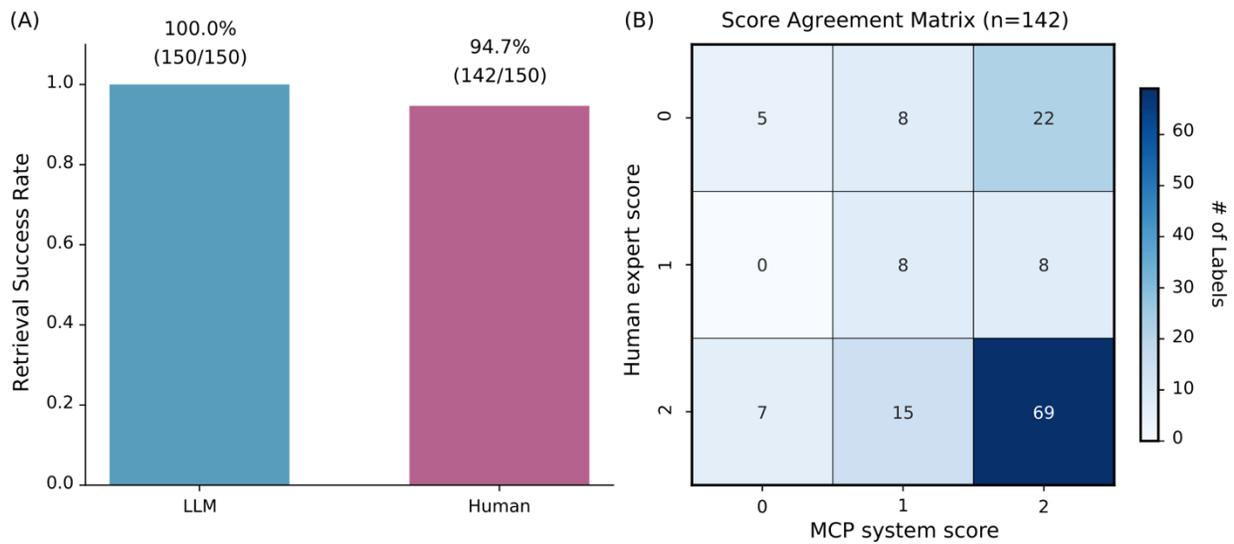

Figure 3. Comparative performance of MCP system versus human experts.

# Discussion

In this study, we introduced a lightweight, training-free system for terminology mapping using the MCP framework to tackle a critical bottleneck in clinical data analysis and research. The system demonstrates that LLMs, when properly guided by structured prompting and access to curated resources, can effectively eliminate hallucination while yielding interpretable mappings to OMOP concepts. Notably, this was achieved without any precomputed embeddings, model fine-tuning, or complex infrastructure setups. Through automating the mapping process with enhanced efficiency and accuracy, our system can significantly mitigate the labor-intensive process while maintaining clinical-grade reliability.

### *The MCP framework alleviates LLM reliability concerns in healthcare*

Beyond operational efficiency, our approach also addresses fundamental concerns about LLM hallucinations in medical applications. Researchers in the healthcare domain have expressed hesitation in adopting LLMs due to their persistent issues with factuality and fabricated outputs. [16] The caution is well-founded, as hallucinations represent an intrinsic limitation of all LLMs, with errors leading to inaccurate fact recording, delayed diagnosis, and patient anxiety. For example, a recent study revealed 1.47% hallucination rates in medical text summarization, with nearly half classified as major errors that could impact patient diagnosis and management if undetected.[17]

Adopting the MCP framework addresses the concerns by providing verified terminology mapping through curated knowledge and vocabulary databases. We demonstrated the success of the MCP framework in eliminating hallucination in concept mapping tasks by ensuring every returned concept ID corresponds to an authenticated OMOP vocabulary entry.

### *Clinical deployment and practical advantages*

Healthcare institutions frequently lack proper infrastructure required for LLM model training and inference. Maintaining in-house LLM implementation tends to be resource-intensive, requiring substantial investment in hardware, software, and skilled personnel. This poses significant barriers to widespread adoption, especially in low-resource healthcare settings.[18]

By contrast, our system can be readily integrated into LLMs that support MCP (e.g., Claude Desktop, Cursor, etc.) or deployed with any LLMs that have tool calling and reasoning capabilities. This eliminates the need for specialized hardware or model training pipelines. It also enables immediate deployment with processing times averaging 6.2 seconds per keyword, which is dramatically faster than manual expert mapping. The absence of infrastructure requirements makes the system more accessible to healthcare organizations of all sizes, from small clinics to large hospital systems.

Furthermore, with our system, users can interact with the LLM in natural language just like they would with any LLM. This conversational interface enables seamless integration with existing expert workflows. For examples, users can provide specific custom instructions such as "prioritize LOINC for laboratory values" or set preferences for custom mapping scenarios. This collaborative, "human-in-the-loop" approach ensures that clinical experts remain engaged in the process, which is a critical requirement for safe AI deployment in healthcare.

### *Contrast with existing mapping solutions*

Many automated mapping solutions have been proposed. For example, OHDSI's Usagi is an interactive tool widely used to map EHR data elements to standardized OMOP concepts based on TF-IDF algorithms.[19] However, Usagi achieved only 90.0% accuracy for common medications and 70.0% for random medications in recent evaluations, requiring extensive post-hoc manual review by domain experts.[20] Additionally, users need to download vocabulary files from Athena and create local Java indexes.

Recent deep learning approaches have shown improved accuracy but introduced deployment barriers. TOKI uses RNN and FastText embeddings and showed modest improvement over Usagi, but researchers had to manually verify 83,000 for training, and it only works for diagnosis conditions.[21] More recent transformer-based approaches, including specialized models like BioLORD-drug and mpnet-drug, achieved higher accuracies ranging from 83-96.5% but the tool requires substantial training datasets and computational infrastructure for model training and inference.[11,12]

More recent LLM-based approaches face similar limitations despite technological advances. One study utilized both RAG and LLM agentic approaches to evaluate vocabulary mapping between Brazilian SIGTAP terminology and OMOP CDM concepts.[13] The RAG pipeline employed textual embeddings to project terms into a shared vector space, followed by LLM-aided term selection, achieving F1 scores of 0.684 for procedures and 0.846 for medicines. However, these approaches were tested on a limited Brazilian healthcare dataset across only two subsets (procedures and medicines) and required multi-stage infrastructure setups such as embedding models and vector storages.

Similarly, another LLM-based mapping automation tool called Llettuce system demonstrated improved performance over traditional approaches, achieving 54% accuracy compared to Usagi's 44% in mapping informal medication names to OMOP concepts.[14] Still, Llettuce needs vector databases and local LLM deployment with processing times averaging 8.7 seconds per concept[14]

Overall, our approach takes a fundamentally different approach. Instead of focusing solely on improving accuracy metrics, our framework tackles deployment barriers that prevented clinical adoption. This represents a paradigm shift toward safety-focused medical AI that deliver both performance and production readiness without the infrastructure complexity.

### *Positioning Within the Emerging MCP Healthcare Ecosystem*

A recent work has also explored MCP applications in clinical decision support, with frameworks integrating LLMs and HL7 Fast Healthcare Interoperability Resources (FHIR) for real-time patient data analysis and clinical decision support.[22] These implementations leverage MCP's systematic approach to accessing diverse FHIR resources, demonstrating the protocol's viability in healthcare AI applications. Our mapping tool could work synergistically with these systems, as standardized concept identification serves as a foundational layer for downstream clinical analytics.

### *Implications for Medical AI Safety and Regulation*

Our findings contribute to the ongoing challenge of deploying LLMs safely in healthcare settings, where hallucination concerns continue to limit adoption.[16,17] By systemically eliminating hallucination through architectural design, the system offer a practical path toward reliable medical AI workflows. In a broader context, the underlying MCP approach can be utilized to enhance safety and reliability across various clinical applications beyond terminology mapping.

# Acknowledgement

This work was supported by the Cancer Prevention and Research Institute of Texas (CPRIT) under award RR230020.

# Conflict of Interest

The authors declare no conflict of interest

# Author Contributions:

J.A. conceived the study, implemented the methodology and software, conducted the analysis, and wrote the manuscript. N.W., H.J., and Z.Y. conducted independent evaluation and validation of system performance and analyzed data. A.W. and S.F. contributed to data curation, evaluation design, conceptual development, and manuscript revision. H.L. supervised all aspects of the research.

# Code Availability

The source code is available at https://github.com/OHNLP/omop_mcp

# References

1 Wang L, Wen A, Fu S, Ruan X, Huang M, Li R *et al.* A scoping review of OMOP CDM adoption for cancer research using real world data. *npj Digit Med* 2025; **8**: 189.

2 Juhn Y, Liu H. Artificial intelligence approaches using natural language processing to advance EHR-based clinical research. *J Allergy Clin Immunol* 2020; **145**: 463–469.

3 Mahadik S, Sen P, Shah EJ. Harnessing digital health technologies and real-world evidence to enhance clinical research and patient outcomes. *Digit Heal* 2025; **11**: 20552076251362097.

4 Reich C, Ostropolets A, Ryan P, Rijnbeek P, Schuemie M, Davydov A *et al.* OHDSI Standardized Vocabularies—a large-scale centralized reference ontology for international data harmonization. *J Am Méd Inform Assoc* 2024; **31**: 583–590.

5 Reinecke I, Zoch M, Reich C, Sedlmayr M, Bathelt F. The Usage of OHDSI OMOP - A Scoping Review. *Stud Heal Technol Inform* 2021; **283**: 95–103.

6 Voss EA, Makadia R, Matcho A, Ma Q, Knoll C, Schuemie M *et al.* Feasibility and utility of applications of the common data model to multiple, disparate observational health databases. *J Am Méd Inform Assoc* 2015; **22**: 553–564.

7 Overhage JM, Ryan PB, Reich CG, Hartzema AG, Stang PE. Validation of a common data model for active safety surveillance research. *J Am Méd Inform Assoc* 2012; **19**: 54–60.

8 Quon JC, Long CP, Halfpenny W, Chuang A, Cai CX, Baxter SL *et al.* Implementing a Common Data Model in Ophthalmology: Mapping Structured Electronic Health Record Ophthalmic Examination Data to Standard Vocabularies. *Ophthalmol Sci* 2025; **5**: 100666.

9 Soroush A, Glicksberg BS, Zimlichman E, Barash Y, Freeman R, Charney AW *et al.* Large Language Models Are Poor Medical Coders — Benchmarking of Medical Code Querying. *NEJM AI* 2024; **1**. doi:10.1056/aidbp2300040.

10 Anthropic. Model Context Protocol. https://modelcontextprotocol.io/.

11 Remy F, Demuynck K, Demeester T. BioLORD: Learning Ontological Representations from Definitions for Biomedical Concepts and their Textual Descriptions. *Find Assoc Comput Linguistics: EMNLP 2022* 2022; : 1454–1465.

12 Song K, Tan X, Qin T, Lu J, Liu T-Y. MPNet: Masked and Permuted Pre-training for Language Understanding. *arXiv* 2020. doi:10.48550/arxiv.2004.09297.

13 Vanzin VJ de B, Moreira D de A, Marcacini RM. LLM-based approaches for automated vocabulary mapping between SIGTAP and OMOP CDM concepts. *Artif Intell Med* 2025; **168**: 103204.

14 Mitchell-White J, Omdivar R, Urwin E, Sivakumar K, Li R, Rae A *et al.* Llettuce: An Open Source Natural Language Processing Tool for the Translation of Medical Terms into Uniform Clinical Encoding. *arXiv* 2024. doi:10.48550/arxiv.2410.09076.

15 FastMCP. https://gofastmcp.com.

16 Ahmad MA, Yaramis I, Roy TD. Creating Trustworthy LLMs: Dealing with Hallucinations in Healthcare AI. *arXiv* 2023. doi:10.48550/arxiv.2311.01463.

17 Asgari E, Montaña-Brown N, Dubois M, Khalil S, Balloch J, Yeung JA *et al.* A framework to assess clinical safety and hallucination rates of LLMs for medical text summarisation. *NPJ Digit Med* 2025; **8**: 274.

18 Dennstädt F, Hastings J, Putora PM, Schmerder M, Cihoric N. Implementing large language models in healthcare while balancing control, collaboration, costs and security. *npj Digit Med* 2025; **8**: 143.

19 OHDSI. OHDSI Usagi. https://ohdsi.github.io/Usagi/.

20 Zhou X, Dhingra LS, Aminorroaya A, Adejumo P, Khera R. A Novel Sentence Transformer-based Natural Language Processing Approach for Schema Mapping of Electronic Health Records to the OMOP Common Data Model. *AMIA Annu Symp Proc AMIA Symp* 2025; **2024**: 1332–1339.

21 Kang B, Yoon J, Kim HY, Jo SJ, Lee Y, Kam HJ. Deep-learning-based automated terminology mapping in OMOP-CDM. *J Am Méd Inform Assoc* 2021; **28**: 1489–1496.

22 Ehtesham A, Singh A, Kumar S. Enhancing Clinical Decision Support and EHR Insights through LLMs and the Model Context Protocol: An Open-Source MCP-FHIR Framework. *arXiv* 2025. doi:10.48550/arxiv.2506.13800.